\documentclass[10pt,twocolumn,letterpaper]{article}

\usepackage[pagenumbers]{cvpr}
\usepackage[T1]{fontenc}
\usepackage{microtype}
\usepackage{threeparttable}
\usepackage{xurl}

\definecolor{cvprblue}{rgb}{0.21,0.49,0.74}
\usepackage[breaklinks,colorlinks,allcolors=cvprblue]{hyperref}
\hypersetup{
  pdftitle={Learning to Track Cells from Sparse Annotations in 3D Time-Lapse Microscopy},
  pdfauthor={Yusuke Uchida}
}

\title{Learning to Track Cells from Sparse Annotations in 3D Time-Lapse Microscopy}
\author{Yusuke Uchida\\
GO Drive Inc.\\
Tokyo, Japan}

\begin{document}
\maketitle
\begin{abstract}

Reconstructing cell lineages from 3D time-lapse microscopy requires distinguishing cell motion from division and inferring temporally consistent connections. When only a subset of cells is annotated, however, training procedures that treat unannotated regions as background or incorrect correspondences can be problematic. We propose a cell tracking pipeline that integrates detection from sparse annotations, calibration of candidate connections using image features, and joint selection of lineage graph events. For detection, we combine the outputs of 2.5D and 3D models trained to suppress erroneous background supervision at unannotated cells. We then calibrate ordinary link and division probabilities using motion features from dense flow, correspondence features from a learned cell-matching model, and features from models that identify cells immediately before division, dividing parents, and daughters immediately after division. Ordinary links, divisions represented as daughter pairs, and node inclusion are selected jointly under a surrogate objective derived from the evaluation metric, followed by gap filling and temporal coordinate refinement. The pipeline achieves an overall score of 0.9762 in 5-fold cross-validation on 199 videos from 2 zebrafish embryos. With candidates and calibrated values held fixed, joint selection outperforms a sequential method that allows replacement of existing divisions and reoptimization of ordinary links by 0.0105. The improvement arises primarily from division recovery.

\end{abstract}

\section{Introduction}
\label{sec:intro}

Tracking cell motion and division during development is essential for understanding how biological tissues form. Light-sheet microscopy of zebrafish embryos has enabled large-scale reconstruction of nuclear positions, movements, and divisions throughout an embryo~\cite{keller2008}. Nevertheless, identifying individual cells and manually recording temporal correspondences and parent--daughter relationships in large image collections requires substantial effort. Automated tracking also remains challenging in densely populated tissues, where cells have similar appearances and images contain noise and intensity variations. The Cell Tracking Challenge has advanced the comparison and generalization assessment of cell segmentation and tracking methods through diverse 2D and 3D time-lapse datasets and a common evaluation protocol~\cite{ctc2023}.

Several approaches learn from limited annotations. 3D U-Net demonstrated volumetric segmentation trained from annotations on a subset of image slices~\cite{unet3d}. For cell tracking, ELEPHANT combines a small number of nuclear annotations with interactive corrections to incrementally train 3D nuclear detection and optical-flow-based tracking. During detector training, background labels are supplied manually or through intensity thresholding in addition to nuclear annotations, and voxels with neither label are excluded from the supervised loss. Its supervision is progressively expanded through cycles of annotation, prediction, and correction~\cite{elephant}.

Temporal correspondence learning has also been addressed in different ways. EmbedTrack jointly performs segmentation and tracking in 2D images by learning offsets from each pixel to the corresponding cell centers in the current and preceding frames~\cite{embedtrack}. Trackastra uses a Transformer to model features such as cell positions and shapes within a temporal window and learns associations that include the one-to-many correspondences arising from division~\cite{trackastra}. Building on the use of motion estimation and context among cells, we evaluate correspondences using motion-compensated spatial relationships and learned appearance features.

Graph optimization has been used to enforce consistency among candidates during lineage reconstruction. Conservation Tracking handles divisions, oversegmentation, and undersegmentation through a probabilistic graphical model with constraints on the consistency of cell counts within detections~\cite{conservation}. Ultrack retains multiple segmentation hypotheses and selects candidates and connections according to temporal consistency, allowing segmentation ambiguity to be resolved during tracking~\cite{ultrack}. We likewise combine image-derived scores with structural constraints, addressing competition between ordinary links and divisions in a graph whose nodes represent cell centers.

Linajea is closely related: it learns cell centers and movement from sparse point annotations and selects lineage graphs using integer linear programming~\cite{linajea}. Its detection loss is restricted to neighborhoods of annotated centers. Hirsch et al. extend this framework with parent, daughter, and continuation-state scores and consistency constraints~\cite{linajea-states}. These studies establish sparse supervision and division-aware global selection as existing approaches. Our design retains background supervision outside regions likely to contain unannotated cells and combines motion, matching, and division-stage features to calibrate daughter-pair events. We select these events together with supporting ordinary links using a metric-derived objective, and evaluate this combination through comparisons with shared candidates and calibrated values.

The Biohub - Cell Tracking During Development competition was organized to develop cell tracking methods for this task.\footnote{\href{https://www.kaggle.com/competitions/biohub-cell-tracking-during-development}{Biohub -- Cell Tracking During Development}, Kaggle, 2026.} The inputs are 3D time-lapse images of fluorescently labeled zebrafish embryos. The prediction targets are cell center positions at each time point and temporal connections describing cell continuation and division. Predictions are represented as a lineage graph with cell detections as nodes and temporal connections as directed edges. The task requires center detection and lineage reconstruction rather than cell segmentation.

A central challenge in this dataset is that annotations cover only a subset of cells. Across 199 videos from 2 embryos, the pooled ratio of annotated nodes to the estimated total number of nodes is approximately 2.8\%. This ratio concerns cell--time pairs and does not directly indicate the fraction of individual cells tracked within an embryo. Treating every unannotated location as background may suppress the detection of real cells. Moreover, there are only 151 annotated divisions, counted across all crops, requiring division morphology and parent--daughter relationships to be learned from few positive examples.

Proximity-based association becomes ambiguous when cell motion, division, and temporary missed detections occur together. Even if a division-specific appearance is identified, the lineage cannot be reconstructed correctly when its daughters are assigned to other tracks. Local correspondence assessment using image features therefore needs to be integrated with structural constraints on parent--daughter relationships.

We propose a tracking method that integrates detection from sparse annotations, calibration of candidate connections using motion, appearance, and division features, and joint selection of lineage graph events. Because detection quality affects both downstream correspondence estimation and division recovery, we design detection and lineage selection as a unified pipeline. Our contributions are threefold.

First, to improve detection under sparse annotation, we combine training that suppresses erroneous background supervision at unannotated cells with an ensemble of 2.5D and 3D models. Heatmaps from a 2.5D U-Net initialized with 2D pretrained weights and a SegResNet using 3D convolutions are combined to produce a common set of cell candidates for downstream tracking.

Second, we integrate dense-flow motion features, matcher correspondence features, and image features that identify stages before and after division to calibrate ordinary link and daughter-pair probabilities. Let \(t\) denote the time of the dividing parent. We identify its predecessor at \(t-1\), the parent at \(t\), and its daughters at \(t+1\) as pre, parent, and post, respectively. By estimating separately the probability that a candidate is evaluated and the probability that it is correct conditional on evaluation, we translate motion, appearance, and division-stage evidence into the expected contribution of a candidate connection.

Third, we construct a surrogate objective from the edge term with its node-count adjustment and the division term, and jointly select ordinary links, divisions represented as daughter pairs, and node inclusion under connection-capacity and temporal-support constraints. A division and the preceding and following ordinary links needed to support it are selected together, ensuring consistency among competing connections.

We evaluate the complete pipeline and use ablations with shared detection candidates to assess calibration features, calibrators, information from before and after division, and the selection strategy. We also compare division input alignment with the number of inputs and model capacity held fixed, and evaluate transfer of the downstream models when the detector is replaced by a publicly available model.

Code for training, inference, and evaluation is publicly available at \url{https://github.com/yu4u/kaggle-biohub-3rd}.

\section{Proposed Approach}
\label{sec:approach}

Let \(I_t\) denote the input image at time \(t\), and \(V_t\) the set of candidate cells. The proposed method detects \(V_t\), assigns correspondence and division scores to candidates in adjacent frames, and determines connections under constraints on the entire graph. Unless otherwise specified, distances and displacements are expressed in physical coordinates in \ensuremath{\mu}m. We refer to annotated cell positions and connections as ground truth (GT).

The detector combines heatmaps from 2.5D and 3D models and outputs a position and confidence for each candidate. Dense flow estimates motion between adjacent frames, and cell matching computes correspondence scores from motion-compensated spatial relationships and image features. The same flow is also used to align the inputs of the division-parent model. Scores for pre, parent, and post, corresponding to cells immediately before division, dividing parents, and daughters immediately after division, are used together with motion and correspondence features to calibrate ordinary links and division candidates.

\begin{figure*}[t]
\centering
\includegraphics[width=\textwidth]{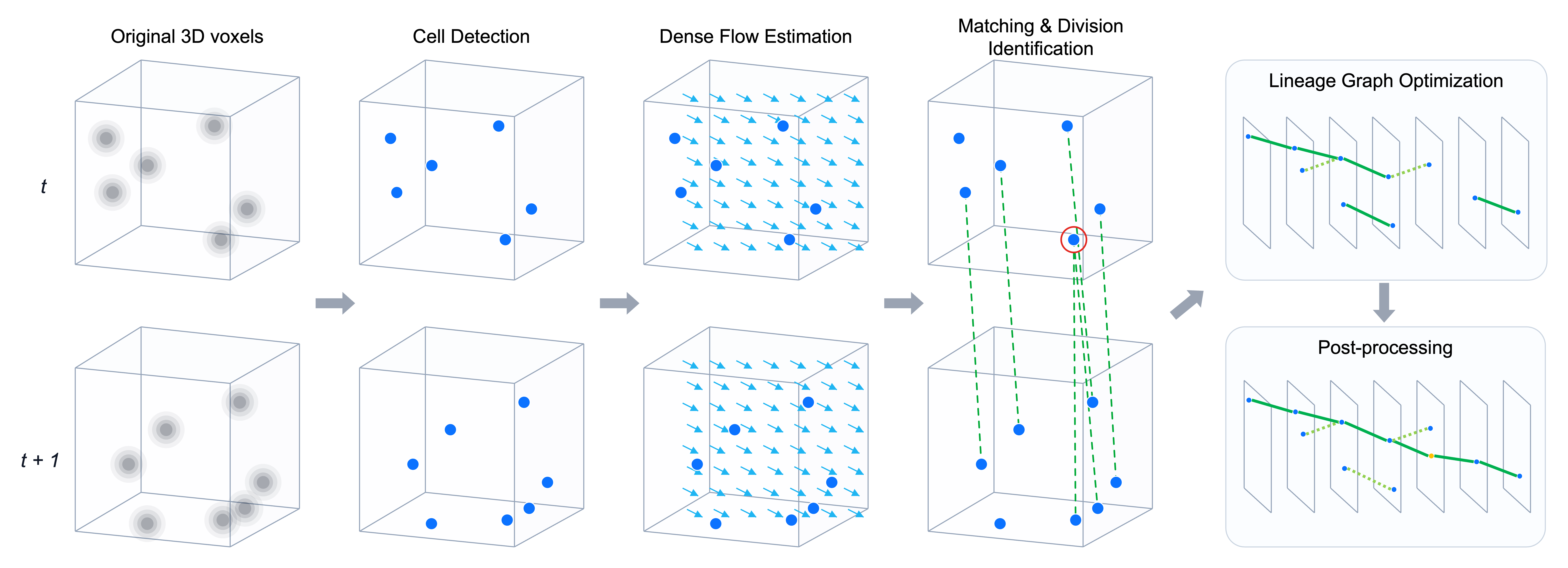}
\caption{\textbf{Inference pipeline of the proposed method.} The upper and lower rows on the left represent 3D images at adjacent times \(t\) and \(t+1\). Cell detection, dense flow estimation, matching, and division identification produce candidate correspondences and divisions. Blue points denote detections, cyan arrows denote flow, green dashed lines denote candidate temporal correspondences, and the red circle denotes a candidate dividing parent. The upper-right panel illustrates lineage graph optimization using calibrated scores and structural constraints; the lower-right panel illustrates gap filling, removal of short components, and coordinate refinement. Images, detections, and lineages are schematic.}
\label{fig:pipeline}
\end{figure*}

Lineage graph optimization integrates these scores and selects nodes, ordinary links, and divisions, followed by gap filling and coordinate refinement. Image models and calibrators are trained in stages, and their predictions are integrated through graph optimization. This design allows downstream components to be compared with detection candidates held fixed. We address offline tracking, where the entire video is available; division identification and coordinate refinement also use future frames.
Figure~\ref{fig:pipeline} summarizes the inference pipeline.

\subsection{Backbone Architecture for 3D Volumes}
\label{sec:backbone}

We primarily use a 2.5D U-Net to extract features from 3D images. This architecture integrates information along the depth axis while using an encoder pretrained on 2D images. The basic design follows 2.5D U-Net with Depth Reduction~\cite{unet25d} by Uchida and Fukui.

At each encoder stage, z slices are folded into the batch dimension for 2D convolution. The stage output is restored to its 3D arrangement, reduced in depth by average pooling along z, and passed to the next stage. The decoder receives 3D features through skip connections and restores spatial resolution using upsampling and 3D convolutions. This architecture aggregates depth context hierarchically while retaining pretrained 2D encoder weights for initialization.

For the 2.5D detection and matching models, each input slice is stacked along the channel dimension with its neighboring z slices, yielding 3 channels for the 2D encoder. This exposes the earliest feature extraction stage to local depth information. These channels represent spatial neighbors and are distinct from the temporal inputs used in Section~\ref{sec:division}. In the division models, a shared encoder processes each single-channel image separately before temporal information is integrated.

We use EfficientNetV2~\cite{efficientnetv2} and EfficientNet~\cite{efficientnet} as 2D encoders. Detection uses EfficientNetV2-L and EfficientNet-B7, whereas matching and division identification use EfficientNetV2-S. The matching model outputs intermediate decoder features, and the detection models output single-channel heatmaps. The architecture is shared across tasks, but the weights are trained independently.

For detection, we also use the fully 3D SegResNet~\cite{segresnet} implemented in MONAI~\cite{monai}. It has a residual encoder--decoder built from 3D convolutions, 32 initial channels, 4 encoder stages with 1 residual block per stage, and Group Normalization. Its heatmap is included in the ensemble because its feature extraction differs from that of the 2.5D models. The flow model is a separate, compact 3D network described in Section~\ref{sec:flow}.

\subsection{Cell Detection}
\label{sec:detection}

We formulate cell center detection as 3D heatmap regression. Image intensities are normalized linearly using the 0.1st and 99.9th percentiles of each video, and negative values are clipped to 0. The target heatmap is constructed by placing a Gaussian with standard deviation \(\sigma\) at each annotated point \(\mathbf{c}_k\), taking the maximum where Gaussians overlap:

\begin{equation}
H(\mathbf{x})=\max_k\exp\left(-\frac{\|\mathbf{x}-\mathbf{c}_k\|_2^2}{2\sigma^2}\right).
\label{eq:1}
\end{equation}

Coordinates are expressed in physical units, with \(\sigma=2.0\) \ensuremath{\mu}m. In the discrete implementation, each Gaussian is evaluated inside an axis-aligned box extending \(\lceil3\sigma/h_d\rceil\) voxels from its center along axis \(d\), where \(h_d\) is the voxel spacing, and is zero outside that box. During training, 3D crops are sampled around annotated points and augmented with flips, rotations in the xy plane, and intensity changes. Target heatmaps are generated from the annotation coordinates after the geometric transformations.

To accommodate sparse annotations, we exclude regions likely to contain unannotated cells from the training loss. Cell candidates are extracted using a low-threshold Difference of Gaussians (DoG), and regions within 6 \ensuremath{\mu}m of candidates not associated with annotations are masked. Candidates within 2 \ensuremath{\mu}m of an annotation are treated as duplicate detections of the same cell and are not used as mask centers. Annotated cores with \(H\geq\exp(-1.5^2/2)\) remain supervised even when they overlap a mask. This suppresses erroneous background supervision at unannotated cells while allowing regions away from candidates to provide background supervision.

Because positive regions occupy a small fraction of the volume, errors are normalized separately over positive and background regions. The main detection loss is

\begin{equation}
\mathcal{L}_{\mathrm{det}}
= \operatorname{mean}_{\Omega_{\mathrm{pos}}}(\hat{H}-H)^2
+ \lambda_{\mathrm{bg}}\operatorname{mean}_{\Omega_{\mathrm{bg}}}(\hat{H}-H)^2,
\label{eq:2}
\end{equation}

where \(H\) and \(\hat{H}\) are the target and predicted heatmaps. Among unmasked voxels, \(\Omega_{\mathrm{pos}}\) contains those with \(H\geq\exp(-9/2)\), and \(\Omega_{\mathrm{bg}}\) contains those below this threshold. An empty region contributes zero loss. We set \(\lambda_{\mathrm{bg}}=0.5\). Separate normalization prevents the large number of background voxels from overwhelming the positive training signal. DoG is used only to identify unannotated regions to exclude during training; inference detections are generated from the learned heatmaps.

At inference, heatmaps from EfficientNetV2-L, SegResNet, and EfficientNet-B7 are combined with weights of 0.3, 0.3, and 0.4, respectively. We use 2-view test-time augmentation (TTA) with the original image and an x-axis flip, undo the flip, and average the heatmaps on the same spatial grid.

Local maxima are extracted by non-maximum suppression, and points with confidence at least 0.2 are retained as cell candidates. Their coordinates and confidence values are passed to downstream matching and graph optimization.

\subsection{Dense Flow Estimation}
\label{sec:flow}

Association based only on distance, without accounting for cell motion, can fail under collective movement and local deformation. We therefore estimate dense flow between adjacent 3D images and predict the location of each detected cell in the next frame.

Images are downsampled by factors of \((z,y,x)=(1,4,4)\), yielding an isotropic grid with 1.625 \ensuremath{\mu}m spacing along each axis. A shared 3D encoder processes the two images, and local correlations are computed between features at half resolution. Correlations over candidate displacements are normalized with a softmax, and their expected displacement provides the initial flow. This flow is interpolated to the input resolution and refined by a residual predicted by a decoder that combines deeper features from both images.

The flow model has 3 encoder stages with 32, 64, and 128 channels. Each block consists of 3D convolutions and GELU activations without normalization layers. The output is a displacement vector with 3 components in the \((z,y,x)\) directions at each voxel. This allows motion to vary spatially rather than restricting the entire image to a single translation.

The training objective combines self-supervised losses on real image pairs with a supervised loss on synthetic deformations. For real pairs, we use image intensity error after warping, error in the census descriptor of local contrast, forward--backward cycle consistency, spatial flow smoothness, and feature consistency under warping. Weighting by fluorescence signal and local contrast reduces the tendency toward a zero-flow solution that explains only the dominant background.

Synthetic pairs are generated by applying known translations and smooth local deformations to real images. Supervision with known displacements complements motion learning in regions where intensity consistency alone is ambiguous. The 199 training videos are shuffled with random seed 42 and split into 179 videos for flow training and 20 for validation, independently of the tracking CV folds. Real training pairs and the source images for synthetic pairs come from the 179-video subset. GT cell links are used only in the 20-video validation subset to assess endpoint error for model selection. The selected checkpoint is epoch 48, which also has the lowest validation endpoint error in the saved training history. The same flow weights are used across all CV folds and for hidden-test inference, without adaptation to hidden-test images.

For a cell at position \(\mathbf{x}_i\) at time \(t\), let \(\mathbf{f}_t\) denote flow converted to physical units. Its predicted position in the next frame is \(\mathbf{x}_i+\mathbf{f}_t(\mathbf{x}_i)\). The motion-compensated distance to a candidate \(\mathbf{x}_j\) in the next frame is

\begin{equation}
d_{ij}=\|\mathbf{x}_j-\mathbf{x}_i-\mathbf{f}_t(\mathbf{x}_i)\|_2,
\label{eq:3}
\end{equation}

and is used in correspondence scoring and graph construction. Dense flow is computed once for the entire image and reused both to sample point displacements and to align images as described in Section~\ref{sec:division}.

\subsection{Cell Matching}
\label{sec:matching}

Cell matching assigns correspondence scores to candidate cells in adjacent frames. Both frames, downsampled by \((1,4,4)\), are processed by an EfficientNetV2-S 2.5D U-Net trained independently of the detector. Weights are shared between the frames, and decoder features at stride 2 relative to the input grid are sampled at detection positions using trilinear interpolation. We directly use the 64-channel intermediate decoder output, omitting the subsequent higher-resolution decoder and detection head.

Positional encodings are added to each cell feature, and self-attention and cross-attention incorporate context within cell sets and between frames. Learnable role embeddings distinguish the preceding and following frames, encoding temporal direction. Correspondence scores use appearance cosine similarity as a prior score, with a correction derived from the attention features.

Candidate matches are selected from motion-compensated neighborhoods. In addition to candidate scores, the matcher outputs a null score for the absence of a correspondence. Forward and backward conditional probabilities distinguish associations supported in both directions from those with strong support in only one direction. Training uses a 16 \ensuremath{\mu}m neighborhood, while the basic downstream candidate set contains edges among the top 5 matcher candidates whose motion-compensated distance is within 12 \ensuremath{\mu}m.

Training targets are constructed by matching detections to GT points and transferring annotated edges to pairs of detections. Points with annotated GT connections serve as anchors. Cross-entropy is computed in both directions to select the correct match from a softmax over the candidates and the null option. Unannotated cells remain competing candidates, but anchors without annotated connections are not automatically assigned null targets.

A dividing parent has 2 correct daughters. We construct a separate training row for each daughter and exclude the other daughter from the negative candidates in that row. To simulate missed detections, the correct candidate is hidden with a fixed probability and the matcher is trained to select the null option. During training only, missed GT endpoints are added to ensure that positive pairs are available. No GT-based candidates are added during validation or inference.

The matcher provides correspondence probabilities, candidate ranks, margins relative to competing candidates, and support in both temporal directions. These outputs are candidate-level evidence rather than final connections. When multiple cells compete for the same match, or ordinary links compete with divisions, selection is determined by the optimization in Section~\ref{sec:optimization}.

\subsection{Division Identification}
\label{sec:division}

In a single frame, division can be difficult to distinguish from ordinary cells or nearby cell pairs. Temporal image changes provide useful information, but reflect both division-related morphological changes and cell motion. We therefore use images at \(t-1\), \(t\), and \(t+1\), combining raw and motion-compensated images to estimate whether a cell at \(t\) is a parent that divides in the next frame.

\textbf{Flow-based input alignment.} Inputs to the division model are downsampled by \((1,2,2)\). Flow from Section~\ref{sec:flow} is interpolated to this grid, with displacement units converted accordingly. Let \(\mathbf{v}\) denote coordinates on the input grid, and \(\mathbf{u}_t\) the forward flow from \(t\to t+1\) in the same grid units. The next frame aligned to the central frame is

\begin{equation}
\widetilde I_{t+1\to t}(\mathbf{v})
=I_{t+1}\bigl(\mathbf{v}+\mathbf{u}_t(\mathbf{v})\bigr),
\label{eq:4}
\end{equation}

and is obtained by trilinear sampling. The preceding frame is aligned using the stored forward flow as

\begin{equation}
\widetilde I_{t-1\to t}(\mathbf{v})
\approx I_{t-1}\bigl(\mathbf{v}-\mathbf{u}_{t-1}(\mathbf{v})\bigr).
\label{eq:5}
\end{equation}

The latter evaluates the \(t-1\to t\) displacement at destination coordinates and approximates the inverse mapping rather than solving for an exact inverse flow. We use this approximation under the assumption that the displacement field is locally smooth. Grid conversions for images and flow also account for the shift in voxel centers caused by average pooling. At the beginning and end of a video, the central frame is duplicated without accessing flow for nonexistent time points.

The parent model receives the following five images:

\begin{equation}
\begin{aligned}
\mathcal I_t=\bigl[&I_{t-1},\ \widetilde I_{t-1\to t},\ I_t,\\
&\widetilde I_{t+1\to t},\ I_{t+1}\bigr].
\end{aligned}
\label{eq:6}
\end{equation}

The inputs consist of raw images from 3 time points and 2 aligned images. Flow compensation can reduce the effect of movement, but a single displacement cannot fully represent the one-to-two correspondence at division. Including the raw images preserves access to the original appearance and spatial arrangement.

\textbf{Feature integration.} Each of the five images is processed by an EfficientNetV2-S encoder with shared weights. Features from each encoder stage are concatenated along the channel dimension in the input order above and passed to a 3D decoder. The decoder uses Group Normalization, stops at stride 2, and produces a logit map through a 1 \ensuremath{\times} 1 \ensuremath{\times} 1 convolution. Logits are interpolated to the input grid, sampled at detection positions, and passed through a sigmoid to obtain parent scores. The flow model remains fixed, and gradients from the division loss are not propagated to it.

\textbf{Targets and training.} Among detections matched to GT, parents with two children in the next frame are positive examples. Cells with ordinary continuation and daughters immediately after division are negatives. Points whose status cannot be determined, for example because an annotation ends, are excluded. The selected parent model does not include detections unmatched to GT as negatives in its loss. Missed GT parents are added to the candidates during training only.

Binary cross-entropy is averaged separately over the positive set \(\mathcal P\) and negative set \(\mathcal N\):

\begin{equation}
\mathcal L_{\mathrm{division}}
=-\frac{1}{|\mathcal P|}\sum_{i\in\mathcal P}\log p_i
-\frac{1}{|\mathcal N|}\sum_{i\in\mathcal N}\log(1-p_i).
\label{eq:7}
\end{equation}

Positive division frames account for 25\% of training samples, and hard-negative frames immediately before the parent stage or just after daughter appearance account for another 25\%; the remainder is sampled from ordinary frames. Common xy transformations and intensity augmentation are applied to all inputs so that augmentation itself does not create artificial temporal changes. The five-input parent model is trained for 20 epochs using AdamW, an initial learning rate of \(10^{-4}\), 3 warmup epochs, and cosine decay to a minimum learning rate of \(10^{-6}\). Evaluation uses an exponential moving average (EMA) with decay 0.995.

\textbf{Auxiliary scores before and after division.} In addition to the parent model, we train pre and post models independently. Positive pre examples are cells whose sole successor divides at the next time point; positive post examples are daughter cells whose parent divided immediately beforehand. Points lacking the preceding or following annotations needed to determine their status are excluded. The auxiliary models take 3 unaligned frames centered on the time being classified. For example, if a dividing parent is at time \(t\), the pre score of its predecessor uses images at \(t-2,t-1,t\).

Parent, pre, and post scores provide evidence about individual cells at stages surrounding division. Because the loss is normalized separately by class, the sigmoid outputs are not directly interpreted as probabilities reflecting the actual division frequency. Daughter pairs are calibrated using parent scores, daughter candidate positions, correspondence scores, and the surrounding division stages; final selection accounts for competition with ordinary links.

For ordinary links and expanded division candidates, we use scores from a separate parent model with 3 unaligned image inputs. The five-input parent model combining raw and flow-aligned images is used to generate and calibrate the basic division candidates. Parent, pre, and post inference uses 4-way rotation TTA in the xy plane and fp16.

\subsection{Lineage Graph Optimization}
\label{sec:optimization}

We jointly select nodes and temporal connections from the detection, matching, and division identification outputs. If ordinary connections are fixed before divisions are added, candidate daughters may already be assigned to other connections, preventing the correct division from being formed. We address this competition by including ordinary links and divisions in a common optimization problem.

\textbf{Candidate event construction.} Ordinary link candidates are the top 5 matcher candidates within a motion-compensated distance of 12 \ensuremath{\mu}m. Basic division candidates are formed from pairs of candidate edges sharing a parent. Candidate generation requires a score of at least 0.1 from the five-input parent model and checks daughter geometry and support from preceding and following correspondence candidates. We additionally use the unaligned three-input parent model and support from preceding and following links to obtain up to 5 new daughter candidates per parent within a radius of 16 \ensuremath{\mu}m. Pairs consisting of a representative daughter from the basic candidates and a new daughter are added as expanded division candidates. Separate calibrators are used for the basic and expanded candidates.

The parent-score threshold limits the candidates considered computationally; it is not a probability threshold for accepting a division. The optimization can revise a daughter's connection even if the initial graph assigns it to another parent.

\textbf{Score calibration.} Features include motion-compensated distance, forward and backward matcher scores, candidate ranks, margins relative to competing candidates, detection confidence, and parent, pre, and post scores. Pre scores of predecessor cells are aggregated over input candidate edges, not only over the edges ultimately selected. Examples include the maximum product of a predecessor candidate's pre score and matcher probability, the pre score of the highest-ranked candidate, and a probability-weighted mean. This makes temporal evidence from multiple possible lineages available before the graph is selected.

With sparse annotations, candidates fall into three categories: true positive (TP), false positive (FP), and unevaluated. For an ordinary link candidate \(e\), we learn separately the probability of being evaluated, \(a_e\) (Exposure), and the probability of being correct conditional on evaluation, \(q_e\) (Conditional). Their expected contributions are

\begin{equation}
p_e^{\mathrm{TP}}=a_eq_e,\qquad
p_e^{\mathrm{FP}}=a_e(1-q_e).
\label{eq:8}
\end{equation}

The model for \(q_e\) is trained on candidates that can be labeled as TP or FP, without treating unevaluated candidates as negatives. The target for \(a_e\) indicates whether the candidate is evaluated. Regularized logistic regression is used for ordinary links.

Divisions are calibrated similarly. For the conditional probability of a daughter pair, the logit of a linear model with 19 features provides an initial score. LightGBM and XGBoost each learn a nonlinear residual over 100 boosting rounds. Both tree models receive the same 34 features, including daughter-to-daughter distance, distances and angular separation relative to the parent, parent and daughter appearance scores, and support from predecessor and successor candidates. These include 6 motion features, such as the discrepancy between the daughter midpoint and an extrapolation using the parent's preceding velocity, changes in daughter velocities across time, and the predicted change in daughter separation in the next frame. The conditional probabilities from the two models are averaged with equal weights. Exposure is estimated by a separate linear model.

Optimization starts from a feasible solution constructed from ordinary links and basic division candidates. LightGBM division probabilities are used for initialization, while the optimization objective coefficients use probabilities averaged with XGBoost.

Division calibration targets indicate whether the parent and both daughters correspond to the two edges of the same GT division. Calibrators are trained at the candidate level, whereas final tracking performance is measured by the official metric, which accounts for temporal tolerance and local lineage structure.

\textbf{Selection variables and constraints.} Let \(y_i\) denote inclusion of candidate node \(i\), and \(x_e\) selection of event \(e\), with both variables binary. An event is either an ordinary link between one parent and one child or a division connecting one parent to two daughters. A daughter pair is selected as a unit rather than as two independent edges.

Let \(\mathcal E_{\mathrm{in}}(i)\) and \(\mathcal E_{\mathrm{out}}(i)\) denote the sets of events entering and leaving node \(i\), respectively. The basic constraints are

\begin{equation}
\begin{aligned}
\sum_{e\in\mathcal E_{\mathrm{in}}(i)}x_e\leq y_i,\\
\sum_{e\in\mathcal E_{\mathrm{out}}(i)}x_e\leq y_i,\\
x_e,y_i\in\{0,1\}.
\end{aligned}
\label{eq:9}
\end{equation}

A division counts as one outgoing event for the parent and consumes the incoming capacity of each daughter. These constraints prevent a cell from being assigned multiple parents and prevent ordinary links and divisions from simultaneously claiming the same daughter. Time always advances along an edge, so the graph is acyclic. After each solve, node inclusion is reconstructed from the endpoints of selected events. Isolated nodes are therefore removed, including those selected at zero node cost, before node counts and the surrogate are recomputed.

Divisions require local temporal support. Let \(C_{\mathrm{in}},C_{\mathrm{out}}\) be the counts of selected incoming and outgoing ordinary links, and \(D_{\mathrm{in}},D_{\mathrm{out}}\) the corresponding division event counts. We impose

\begin{equation}
\begin{aligned}
D_{\mathrm{out}}(i)\leq C_{\mathrm{in}}(i),\\
D_{\mathrm{in}}(i)\leq C_{\mathrm{out}}(i),\\
D_{\mathrm{in}}(i)+D_{\mathrm{out}}(i)\leq1.
\end{aligned}
\label{eq:10}
\end{equation}

A dividing parent must have an incoming ordinary link, and each daughter must have an outgoing ordinary link. A daughter immediately after division is also prevented from dividing at the next frame boundary. This structural assumption reflects the temporal resolution of the dataset and suppresses false detections in which successive division candidates support one another. Restricting temporal support to ordinary links also excludes relaxed solutions in which fractional division variables mutually provide support. These constraints alone, however, do not guarantee a division TP under the official metric.

\textbf{Metric-derived surrogate objective.} For video \(s\), let \(T_s,F_s\) be the sums of expected edge TP and FP contributions from selected events, \(U_s,V_s\) the sums of expected division TP and FP contributions, and \(N_s\) the number of selected nodes. For a division, the two parent-to-daughter edges contribute their separately calibrated edge TP and FP probabilities, while daughter-pair calibration determines the division TP and FP probabilities. Basic edges use the ordinary-link calibrators; an added edge in an expanded candidate uses separately trained edge calibrators. Appendix~\ref{sec:optimization-details} gives the event contributions and optimization details. Let \(\widehat K_s,\widehat L_s,\widehat M_s\) denote the numbers of annotated edges, annotated divisions, and total cell nodes estimated from training data. We define the surrogate objective as

\begin{equation}
\begin{aligned}
\Phi(x,y)={}&
\frac{\sum_s\max\{0,u_s(N_s)T_s\}}
     {\sum_s(\widehat K_s+F_s)}\\
&+0.1\frac{\sum_s U_s}{\sum_s(\widehat L_s+V_s)}.
\end{aligned}
\label{eq:11}
\end{equation}

The node-count adjustment factor is capped at 1 during optimization:

\begin{equation}
u_s(N_s)=\min\left\{1,\ 1.1-0.1\frac{N_s}{\widehat M_s}\right\}.
\label{eq:12}
\end{equation}

This cap is introduced in the surrogate to suppress rewards for underestimating the node count. Estimates of \(\widehat K_s,\widehat L_s,\widehat M_s\) are derived from ratios to candidate node counts in the training videos, without accessing GT at inference. Shared objective coefficients, or prices, are computed from expected counts across all videos being processed.

Derivatives of the surrogate at the current solution provide TP rewards and FP and node-count costs. Fixing these coefficients makes each video's selection problem linear. We solve the videos individually, aggregate their counts, and reevaluate the surrogate. The solution is updated only if the surrogate improves, for up to 3 iterations.

\textbf{Optimization solver.} At each iteration, HiGHS solves the LP relaxation. An integral solution is accepted directly; otherwise, a MILP is solved with integrality constraints on the division events. When node rewards are nonpositive and division choices are fixed, the remaining ordinary link and node selection problem can exploit a network-flow structure. Separate time limits are set for the LP and MILP, and returned solutions are checked for capacity feasibility and integrality. If no feasible solution is obtained, the incumbent feasible solution is retained.

The surrogate is based on ratios of expected counts. It is not the expected official score itself, because point matching and local division structure introduce dependencies among events. Even if every linearized problem is solved optimally, global optimality of the nonlinear surrogate or the score against unknown GT is not guaranteed. We assess effectiveness by applying the official metric to the graph after post-processing.

\subsection{Post-processing}
\label{sec:postprocessing}

\textbf{Gap filling and removal of short components.} After optimization, we fill track interruptions caused by temporary missed detections. Collective motion at each frame boundary is estimated from the median displacement of selected links. Accumulated motion compensates the positional difference between a track endpoint and a subsequent track start. Connections are attempted in increasing order of the number of missing frames, \(g=1,2,3\), with a distance tolerance of \(3\sqrt{g+1}\) \ensuremath{\mu}m for a gap of \(g\) frames.

For each time point and gap length, endpoints and starts are matched one-to-one using the Hungarian algorithm. An endpoint already used for a connection is not reused, and shorter gaps are resolved first. For each accepted connection, \(g\) nodes are inserted by linear interpolation between the original endpoint coordinates, creating a sequence of edges between adjacent frames. Weakly connected components with fewer than 6 nodes are then removed. Gap filling precedes removal because discarding all short fragments first would also remove fragments that could be connected.

\textbf{Temporal coordinate refinement.} Detection positions fluctuate between frames, so we refine coordinates while holding connectivity fixed. For selected ordinary links, an affine model predicting displacement from the source position is fitted at each frame boundary by least squares, weighted by the calibrated conditional probability of correctness \(q_{ij}\). The resulting predicted displacements \(\mathbf m_{ij}\) are fixed, and refined coordinates \(\mathbf z_i\) are obtained from the quadratic objective

\begin{equation}
\begin{aligned}
\min_{\{\mathbf z_i\}}\quad&
\sum_i\|\mathbf z_i-\mathbf x_i\|_2^2\\
&+\lambda\sum_{(i,j)\in E_{\mathrm{ordinary}}}
q_{ij}\|\mathbf z_j-\mathbf z_i-\mathbf m_{ij}\|_2^2.
\end{aligned}
\label{eq:13}
\end{equation}

The first term penalizes deviations from the original detections, while the second encourages displacements along selected correspondences to follow collective motion. All coordinates have the same anchoring weight, and we use \(\lambda=2\). The affine model is estimated once from positions before refinement, and coordinate refinement is solved as a sparse linear system.

When refining actual detections, dividing parents and their immediate daughters, gap endpoints, and interpolated nodes are fixed to preserve positions at structures that differ from ordinary motion. We then refine only the interpolated gap nodes while keeping actual detections fixed. The source endpoint is advanced through the affine displacement fields estimated from ordinary links before refinement, and the residual between the resulting position and the fixed destination is distributed in proportion to time. This replaces straight-line interpolation with a trajectory following local motion while preserving both endpoints. Gaps retain linear interpolation if displacement fields for the required time points are unavailable.

Coordinate refinement is computed in \ensuremath{\mu}m and rounded to integer voxel coordinates. Changes that move a node outside the image or cause two nodes to occupy the same coordinates at the same time are rejected. Only detection and interpolated-node positions are refined; node counts, time points, and connections remain fixed.

\section{Experimental Results}
\label{sec:results}

\subsection{Experimental Setup}
\label{sec:setup}

The training data comprise 71 videos from embryo 44b6 and 128 videos from embryo 6bba. Training and test data come from different embryos. Each video contains 100 frames of 64 \ensuremath{\times} 256 \ensuremath{\times} 256 voxels. Voxel spacing is 1.625 \ensuremath{\mu}m along z and 0.40625 \ensuremath{\mu}m in the xy plane. Across all videos, annotations contain 133,318 nodes, 128,883 edges, and 151 divisions, counting annotations in overlapping crops separately. The organizer-provided estimates of total node counts sum to 4,725,117. Dividing the annotated node count by this estimate gives the approximately 2.8\% ratio reported in Section~\ref{sec:intro}.

We compare configurations using video-level 5-fold cross-validation (CV). Unless otherwise specified, scores in the main text are CV results aggregated over all 199 videos. Video IDs are sorted within each embryo and assigned to the 5 folds in round-robin order, so every fold contains videos from both embryos. All frames of a video belong to the same fold; training and validation are not split within a video.

The detectors, matcher, and division models generate out-of-fold (OOF) predictions by excluding each video's fold from training. For CV evaluation, each model uses one set of weights corresponding to the video's held-out fold. Candidate features are constructed from these predictions, and calibrators are also trained without the evaluation fold. Prior rates for total cells, annotated edges, and divisions are estimated from the calibrators' training videos. Dense flow uses the separate training and model-selection procedure described in Section~\ref{sec:flow} and does not follow the same 5-fold OOF scheme.

We use stacking, with calibrators trained on OOF image-model outputs, rather than nested cross-validation that retrains all image models for each outer validation fold. Some crops from the same embryo overlap in space and time; the existing split contains 37 overlapping pairs that cross fold boundaries. Shared optimization prices are computed by pooling the expected counts from all 199 OOF predictions, with each video's calibrators and count priors fitted outside its fold. The optimization is therefore performed jointly over the CV pool rather than separately within each evaluation fold.

The cumulative evaluation and ablations share the detection candidates from the ensemble in Section~\ref{sec:detection}. The initial candidate count is 5,320,079 across all 199 videos. Section~\ref{sec:transfer} replaces the detector to evaluate transfer of the downstream models.

Table~\ref{tab:training} lists the main training settings. The 2.5D detectors use pretrained encoders, while the fully 3D detector is initialized randomly. Detector evaluation uses the EMA weights from the final epoch. The five-input parent model also uses EMA. Among checkpoints whose median scores for validation negatives and unmatched detections in the evaluation fold are below 0.05 and 0.15, respectively, we select the checkpoint with the highest AUC computed from labels in that fold. If no checkpoint satisfies these criteria, the final epoch is used. Scores of unmatched detections are used for this selection diagnostic but are not included in the parent model's negative loss.

\begin{table*}[t]
\centering
\begin{threeparttable}
\caption{Main training settings for the image models.}
\label{tab:training}
\normalsize
\setlength{\tabcolsep}{4pt}
\renewcommand{\arraystretch}{1.15}
\begin{tabular}{@{}lccrrr@{}}
\toprule
Model & \shortstack{Input downsampling\\factors (z, y, x)} & \shortstack{Training region\\(downsampled voxels)} & Epochs & \shortstack[r]{Learning\\rate} & EMA \\
\midrule
2.5D detection & (1, 1, 1) & 48 \ensuremath{\times} 192 \ensuremath{\times} 192 & 800 & 1e-3 & 0.995 \\
3D detection & (1, 1, 1) & 48 \ensuremath{\times} 192 \ensuremath{\times} 192 & 400 & 3e-4 & 0.999 \\
Matcher & (1, 4, 4) & 64 \ensuremath{\times} 64 \ensuremath{\times} 64 & 200 & 1e-3 & None \\
Dense flow & (1, 4, 4) & 64 \ensuremath{\times} 64 \ensuremath{\times} 64 & 48 & 5e-4 & None \\
Division parent & (1, 2, 2) & 64 \ensuremath{\times} 128 \ensuremath{\times} 128 & 20 & 1e-4 & 0.995 \\
\bottomrule
\end{tabular}
\end{threeparttable}
\end{table*}

Both conditions in Section~\ref{sec:alignment} use random seed 42 and a batch size of 8.

\subsection{Evaluation Protocol and Overall Performance}
\label{sec:evaluation}

We use the overall score defined by the official evaluation specification.\footnote{Royer Lab, \href{https://github.com/royerlab/kaggle-cell-tracking-competition/blob/main/metrics.md}{Evaluating tracking predictions}, competition evaluation documentation.} It is a weighted sum of edge Jaccard with a node-count adjustment and division Jaccard:

\begin{equation}
S = J_{\mathrm{edge}}^{\mathrm{adjusted}} + 0.1 J_{\mathrm{division}}.
\label{eq:14}
\end{equation}

Predicted and GT nodes are matched one-to-one at each time point within a physical distance of 7 \ensuremath{\mu}m. A predicted edge is a TP when its endpoints correspond to the correct GT edge. An incorrect connection is an FP when it competes with an annotated connection; other connections in unannotated regions are unevaluated. Division evaluation considers both parent identification and the local graph structure connecting the two daughter lineages around the event, allowing a division-time offset of 1 frame before or after GT.

For video \(s\), define \(w_s=\mathrm{TP}_s+\mathrm{FP}_s+\mathrm{FN}_s\) and \(J_s=\mathrm{TP}_s/w_s\) from edge TP, FP, and false negative (FN) counts. Let \(N_s\) be the predicted node count and \(M_s\) the organizer-provided estimate of the total node count. The official adjustment and aggregation across videos are

\begin{equation}
\begin{aligned}
J_s^{\mathrm{adjusted}}
&=\max\left\{0,\ J_s\left(1-0.1\frac{N_s-M_s}{M_s}\right)\right\},\\
J_{\mathrm{edge}}^{\mathrm{adjusted}}
&=\frac{\sum_s w_sJ_s^{\mathrm{adjusted}}}{\sum_s w_s}.
\end{aligned}
\label{eq:15}
\end{equation}

The division term pools division counts across videos:

\begin{equation}
J_{\mathrm{division}}
=\frac{\sum_s\mathrm{TP}^{\mathrm{div}}_s}
{\sum_s(\mathrm{TP}^{\mathrm{div}}_s+\mathrm{FP}^{\mathrm{div}}_s+\mathrm{FN}^{\mathrm{div}}_s)}.
\label{eq:16}
\end{equation}

A Jaccard term is set to 0 when its denominator is 0. Counts are obtained by applying the organizer's evaluation implementation to each video's lineage graph.

The official node-count adjustment factor \(1.1-0.1N_s/M_s\) is not capped at 1 and exceeds 1 when \(N_s<M_s\). In contrast, the surrogate in Section~\ref{sec:optimization} caps the factor at 1 to suppress rewards for underestimating node counts. The official \(M_s\) is accessed only during scoring; optimization uses \(\widehat M_s\), estimated from training videos. All scores reported in the main text are computed using the official evaluation equations above, not the surrogate objective.

We use ``after optimization'' to denote the graph after joint selection of nodes, ordinary links, and divisions, and ``final output'' to denote the graph after gap filling, removal of short components, and temporal coordinate refinement.

Table~\ref{tab:stages} adds processing stages cumulatively while keeping detection candidates fixed. The baseline Hungarian assignment uses a distance threshold of 7 \ensuremath{\mu}m and 2 rounds of collective-motion correction. We then introduce dense-flow motion compensation, restriction to the top 5 matcher candidates with correspondence-probability costs, and division candidates, calibration, and joint optimization. Rows after optimization apply successive post-processing steps to the same optimized solution.

\begin{table*}[t]
\centering
\begin{threeparttable}
\caption{Cumulative CV performance on fixed detection candidates.}
\label{tab:stages}
\normalsize
\setlength{\tabcolsep}{4pt}
\renewcommand{\arraystretch}{1.15}
\begin{tabular}{@{}lrrr@{}}
\toprule
Processing stage (cumulative changes) & \shortstack[r]{Adjusted edge\\Jaccard} & \shortstack[r]{Division\\Jaccard} & \shortstack[r]{Overall\\score} \\
\midrule
Detection + distance-based assignment with collective-motion correction & 0.8906 & 0.0000 & 0.8906 \\
+ Dense-flow motion compensation & 0.9026 & 0.0000 & 0.9026 \\
+ Matcher candidate selection and correspondence scores & 0.9028 & 0.0000 & 0.9028 \\
+ Division candidates, calibration, and joint optimization & 0.9112 & 0.5185 & 0.9630 \\
+ Gap filling & 0.9167 & 0.5185 & 0.9685 \\
+ Removal of weakly connected components with fewer than 6 nodes & 0.9184 & 0.5185 & 0.9703 \\
+ Temporal coordinate refinement (final output) & 0.9235 & 0.5266 & 0.9762 \\
\bottomrule
\end{tabular}
\end{threeparttable}
\end{table*}

The final output has edge TP/FP/FN counts of 122,313/3,657/6,570, division TP/FP/FN counts of 99/37/52, and 5,046,689 predicted nodes. The overall scores by embryo are 0.9558 for 44b6 and 0.9806 for 6bba. Moving from matcher-based assignment to joint optimization changes divisions, calibration, and candidate coverage together. Section~\ref{sec:joint} isolates the selection strategy by holding candidates and calibrated values fixed.

On the hidden test set, a configuration using a 5-fold ensemble and TensorRT fp16 achieved public and private scores of 0.9780 and 0.9671, respectively.

\subsection{Ablation Studies}
\label{sec:ablations}

To evaluate individual components, we share detections, ordinary link and division candidates, constraints, the initial solution, and post-processing across conditions, changing only the features, probability models, or selection strategy under study. For feature ablations, the corresponding inputs are set to zero during training and inference, and the calibrators are retrained. Model outputs used for candidate generation and initialization are retained. These experiments therefore assess calibration and selection on a fixed candidate set, rather than removing an image model from the entire pipeline, including candidate generation. Section~\ref{sec:alignment} retrains the division-parent models to compare input alignment on fixed candidates.

We estimate 95\% confidence intervals (CIs) using a paired cluster bootstrap stratified by embryo, with connected components of overlapping crops as the resampling units (4,000 repetitions, random seed 67). These intervals describe resampling variability with trained models and predictions held fixed; no correction for multiple comparisons is applied.

\subsubsection{Image-Model Features Used for Calibration}
\label{sec:features}

We first remove calibration features derived from flow, the matcher, and the division models separately. The flow ablation removes motion-compensated distance and its transformations. The matcher ablation removes correspondence probabilities, ranks, competitive margins, support, and motion features derived from the matcher. The division ablation removes parent, pre, and post scores and their products and aggregated features. Candidate sets, training labels and weights, priors, and the initial solution are shared, and calibrators are retrained using 5 folds.
Table~\ref{tab:features} summarizes the results.

\begin{table*}[t]
\centering
\begin{threeparttable}
\caption{Calibration-feature ablations with fixed candidates.}
\label{tab:features}
\normalsize
\setlength{\tabcolsep}{4pt}
\renewcommand{\arraystretch}{1.15}
\begin{tabular}{@{}lrrrr@{}}
\toprule
Features used for calibration & \shortstack[r]{Score after\\optimization} & \shortstack[r]{Final adjusted\\edge Jaccard} & \shortstack[r]{Final division\\Jaccard} & \shortstack[r]{Final overall\\score} \\
\midrule
All features & 0.9630 & 0.9235 & 0.5266 & 0.9762 \\
Without flow features & 0.9612 & 0.9207 & 0.5213 & 0.9728 \\
Without matcher features & 0.9530 & 0.9179 & 0.4607 & 0.9640 \\
Without division-model features & 0.9425 & 0.9227 & 0.3285 & 0.9555 \\
\bottomrule
\end{tabular}
\end{threeparttable}
\end{table*}

Every feature removal reduces the score both after optimization and in the final output. Removing division-model features produces the largest final-score decrease, with division Jaccard falling from 0.5266 to 0.3285 while adjusted edge Jaccard changes from 0.9235 to 0.9227. Removing flow features primarily affects the edge term, whereas removing matcher features reduces both terms.

\subsubsection{Joint Selection of Ordinary Links and Divisions}
\label{sec:joint}

We compare joint selection with two sequential strategies under the same calibrated probabilities and objective. The first sequential strategy selects ordinary links first, then adds divisions through local replacements with positive gain, removing ordinary links that compete for the daughters. The second also allows replacement of selected divisions and alternates between local division replacement and reoptimization of ordinary links with division choices fixed. All strategies share constraints and the initial solution. The outer price update runs for at most 3 iterations, and the sequential strategy allows at most 10 alternating updates.
Table~\ref{tab:selection} summarizes the results.

\begin{table*}[t]
\centering
\begin{threeparttable}
\caption{Joint and sequential selection with shared candidates, calibrated values, and initialization.}
\label{tab:selection}
\normalsize
\setlength{\tabcolsep}{4pt}
\renewcommand{\arraystretch}{1.15}
\begin{tabular}{@{}lrrrr@{}}
\toprule
Selection strategy & \shortstack[r]{After\\optimization} & \shortstack[r]{Final\\output} & \shortstack[r]{Final division\\TP/FP/FN} & \shortstack[r]{Final-output difference from\\joint selection [95\% CI]} \\
\midrule
Joint selection & 0.9630 & 0.9762 & 99/37/52 & --- \\
Select ordinary links, then add divisions locally & 0.9535 & 0.9657 & 78/34/73 & \ensuremath{-}0.0105 [\ensuremath{-}0.0156, \ensuremath{-}0.0052] \\
\shortstack[l]{Also allow division replacement\\and ordinary-link reoptimization} & 0.9535 & 0.9657 & 78/34/73 & \ensuremath{-}0.0105 [\ensuremath{-}0.0156, \ensuremath{-}0.0052] \\
\bottomrule
\end{tabular}
\end{threeparttable}
\end{table*}

Joint selection improves the score by 0.0105 over the sequential strategy with alternating optimization, with positive differences in both embryos and all 5 folds. It recovers 21 additional division TPs with 3 additional FPs, reducing division FNs from 73 to 52. The adjusted edge term remains approximately unchanged, so the overall gain is primarily due to division recovery. These are official evaluation counts across crops, which can include overlapping observations of the same biological event. The alternating sequential strategy converges within at most 2 updates and does not reach the iteration limit. The two sequential strategies differ slightly in score but round to the same displayed values and interval bounds.

\begin{figure*}[t]
\centering
\includegraphics[width=\textwidth]{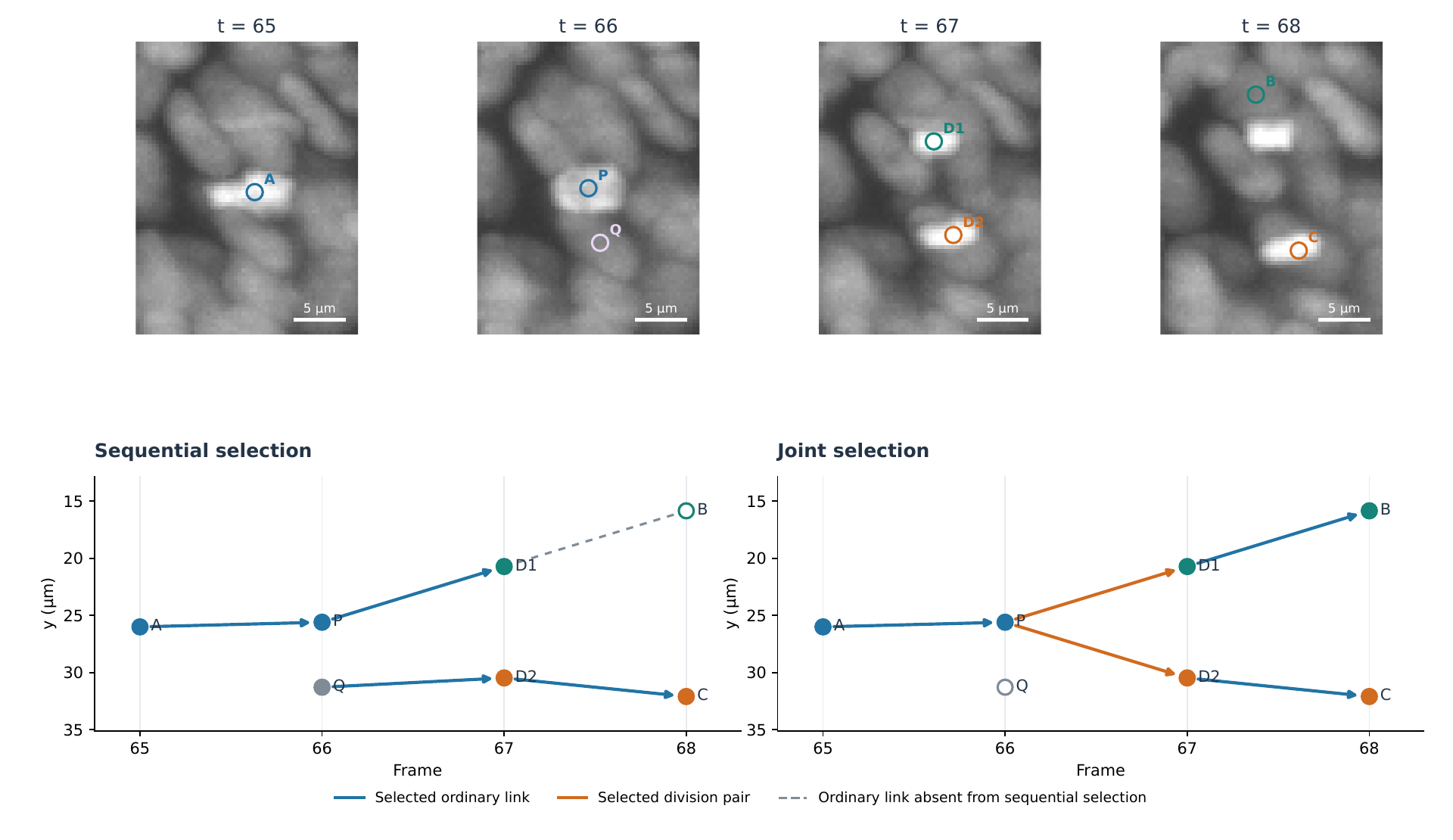}
\caption{\textbf{Joint selection of a division candidate and its supporting links.} Frames 65--68 of video 44b6\_12dfb391 are shown. The upper panels show maximum-intensity projections over a local z range with detections; the lower panels show part of the selections made at the shared first-round prices. The candidate with parent P and daughters D1 and D2 is positive because it corresponds to the two edges of a GT division. The sequential strategy, which allows replacement of existing divisions and reoptimization of ordinary links, selects P\ensuremath{\to}D1 and Q\ensuremath{\to}D2 as ordinary links but lacks a successor link from D1. The joint strategy selects P\ensuremath{\to}\{D1, D2\} as a division and also adds the ordinary link D1\ensuremath{\to}B. Blue lines denote ordinary links, orange lines denote the two edges selected together as a division, and the dashed line denotes an ordinary link absent from the sequential solution. Open circles mark candidates included for comparison. The vertical axis in the lower panels is the y coordinate of each detection; x and z are omitted. This example illustrates candidate selection, not official division TP classification in the final output. Scale bars: 5 \ensuremath{\mu}m.}
\label{fig:joint}
\end{figure*}

Analysis at the shared first-round prices shows that, among 15 positive basic division candidates selected only by the joint strategy, 13 lack an incoming ordinary link to the parent or an outgoing ordinary link from a daughter in the sequential solution. This suggests the value of selecting a division together with the ordinary links supporting it before and after the event, relative to the sequential strategies evaluated here. This analysis uses candidate labels and is distinct from the difference in division TP under the official metric.
Figure~\ref{fig:joint} illustrates a division selected together with its supporting links.

\subsubsection{Exposure, Calibrators, and the Objective}
\label{sec:calibration}

To assess the probability of being evaluated, we hold Conditional fixed and replace Exposure with event-type-specific constants estimated from training videos outside the evaluation fold. We compare three conditions: constant Exposure for ordinary links only, for divisions only, and for both. This preserves the decomposition into TP, FP, and unevaluated probabilities while testing the benefit of learning candidate-specific Exposure.

For division Conditional, we compare a linear model, a model that adds a LightGBM residual to a linear logit, and an ensemble of LightGBM and XGBoost probabilities, using the same 34 features and targets. Missing features in the linear model are imputed with training-fold medians. The tree models use the 19-feature linear model from Section~\ref{sec:optimization} for their initial logits, whereas the standalone linear comparator receives 34 features. Exposure and ordinary link calibration remain fixed.

For the objective, we compare the control against setting the node price in the linearized problem to 0 and against stopping after one solve using prices computed from the initial solution. Setting the node price to 0 retains the node-count adjustment on edges and the surrogate-based decision to accept a new solution.
Table~\ref{tab:calibration} summarizes the results.

\begin{table*}[t]
\centering
\begin{threeparttable}
\caption{Ablations of Exposure, calibrators, and the objective.}
\label{tab:calibration}
\normalsize
\setlength{\tabcolsep}{4pt}
\renewcommand{\arraystretch}{1.15}
\begin{tabular}{@{}lrrr@{}}
\toprule
Condition & \shortstack[r]{After\\optimization} & \shortstack[r]{Final\\output} & \shortstack[r]{Final-output difference\\from control [95\% CI]} \\
\midrule
All features, joint selection, calibration ensemble (control) & 0.9630 & 0.9762 & --- \\
Constant Exposure for ordinary links & 0.9603 & 0.9742 & \ensuremath{-}0.0020 [\ensuremath{-}0.0048, +0.0004] \\
Constant Exposure for divisions & 0.9638 & 0.9761 & 0.0000 [\ensuremath{-}0.0017, +0.0016] \\
Constant Exposure for both & 0.9622 & 0.9753 & \ensuremath{-}0.0009 [\ensuremath{-}0.0035, +0.0016] \\
Linear calibration with 34 features & 0.9598 & 0.9719 & \ensuremath{-}0.0042 [\ensuremath{-}0.0081, \ensuremath{-}0.0005] \\
Linear + LightGBM calibration & 0.9625 & 0.9756 & \ensuremath{-}0.0005 [\ensuremath{-}0.0018, +0.0000] \\
Explicit node price set to 0 & 0.9620 & 0.9763 & +0.0001 [\ensuremath{-}0.0007, +0.0010] \\
Stop after one price-update round & 0.9621 & 0.9741 & \ensuremath{-}0.0021 [\ensuremath{-}0.0046, +0.0002] \\
\bottomrule
\end{tabular}
\begin{tablenotes}[flushleft]
\item[] The upper bound displayed as +0.0000 is positive before rounding.
\end{tablenotes}
\end{threeparttable}
\end{table*}

The calibration ensemble outperforms the 34-feature linear model by 0.0042. The individual benefits of adding a tree model and ensembling could not be established. The ensemble-minus-linear differences are \ensuremath{-}0.0009 for embryo 44b6 and +0.0055 for embryo 6bba, showing that the improvement differs between embryos.

Replacing ordinary link Exposure with a constant changes the score immediately after optimization by \ensuremath{-}0.0027. Setting the node price to 0 changes the score by \ensuremath{-}0.0011 after optimization, with the difference shrinking to approximately 0 after post-processing and reaching +0.0001 in the final output. Allowing up to 3 price-update rounds, with a new solution accepted only if the surrogate improves, gives a point estimate 0.0021 higher than stopping after one round. The final-output comparisons do not establish clear improvements from learned Exposure, the explicit node price, or repeated price updates.

\subsubsection{Auxiliary Information Before and After Division}
\label{sec:auxiliary}

We perform a 2\ensuremath{\times}2 comparison of pre and post features while retaining parent scores. The corresponding features are removed from both ordinary links and divisions, and calibrators are retrained. Removing pre also removes aggregated features derived from predecessor candidates' pre scores.
Table~\ref{tab:auxiliary} summarizes the results.

\begin{table}[t]
\centering
\begin{threeparttable}
\caption{Auxiliary information before and after division.}
\label{tab:auxiliary}
\normalsize
\setlength{\tabcolsep}{4pt}
\renewcommand{\arraystretch}{1.15}
\begin{tabular}{@{}lrr@{}}
\toprule
\shortstack[l]{Division information\\used for calibration} & \shortstack[r]{After\\optimization} & \shortstack[r]{Final\\output} \\
\midrule
Parent only & 0.9557 & 0.9675 \\
Parent + pre & 0.9571 & 0.9695 \\
Parent + post & 0.9632 & 0.9755 \\
Parent + pre + post & 0.9630 & 0.9762 \\
\bottomrule
\end{tabular}
\end{threeparttable}
\end{table}

Averaged over the presence or absence of the other auxiliary feature group, the effects on the final score [95\% CI] are +0.0013 [\ensuremath{-}0.0007, +0.0034] for pre and +0.0073 [+0.0034, +0.0113] for post. The interaction is \ensuremath{-}0.0013 [\ensuremath{-}0.0055, +0.0025]. This comparison supports an average benefit from post, while the effects of pre and the interaction remain unclear.

\subsubsection{Flow Alignment with a Matched Number of Inputs}
\label{sec:alignment}

We compare the presence and absence of alignment while holding the division-parent model's input count, temporal window, and capacity fixed. The control uses raw and flow-aligned images. The ablation replaces each aligned image with a duplicate of the corresponding raw image, yielding \([I_{t-1}, I_{t-1}, I_t, I_{t+1}, I_{t+1}]\). Both conditions use 3 time points, 5 inputs, and 39,382,145 parameters. Training settings and the number of training examples presented per epoch are matched, and each fold is trained for 20 epochs.

Both conditions are trained using the OOF outputs of the 3-detector ensemble in Section~\ref{sec:detection}. Parent scores for basic division candidates are replaced with each model's predictions, and Conditional and Exposure are recalibrated. The candidate set, parent scores used to generate it, parent features for expanded candidates, ordinary links, pre and post scores, and the initial solution are held fixed. The comparison concerns the difference between two models trained under matched conditions.
Table~\ref{tab:alignment} summarizes the results.

\begin{table*}[t]
\centering
\begin{threeparttable}
\caption{Input alignment with matched model capacity and training settings.}
\label{tab:alignment}
\normalsize
\setlength{\tabcolsep}{4pt}
\renewcommand{\arraystretch}{1.15}
\begin{tabular}{@{}lrrr@{}}
\toprule
Division-parent inputs & \shortstack[r]{After\\optimization} & \shortstack[r]{Final\\output} & \shortstack[r]{Final division\\TP/FP/FN} \\
\midrule
Five inputs using raw and flow-aligned images (retrained control) & 0.9630 & 0.9752 & 96/34/55 \\
Five inputs using duplicated raw images & 0.9608 & 0.9730 & 95/41/56 \\
\bottomrule
\end{tabular}
\begin{tablenotes}[flushleft]
\item[] Both rows use newly trained division-parent models; the aligned control is therefore distinct from the main pipeline model used in Sections~\ref{sec:evaluation}--\ref{sec:auxiliary}.
\end{tablenotes}
\end{threeparttable}
\end{table*}

\begin{figure*}[t]
\centering
\includegraphics[width=\textwidth]{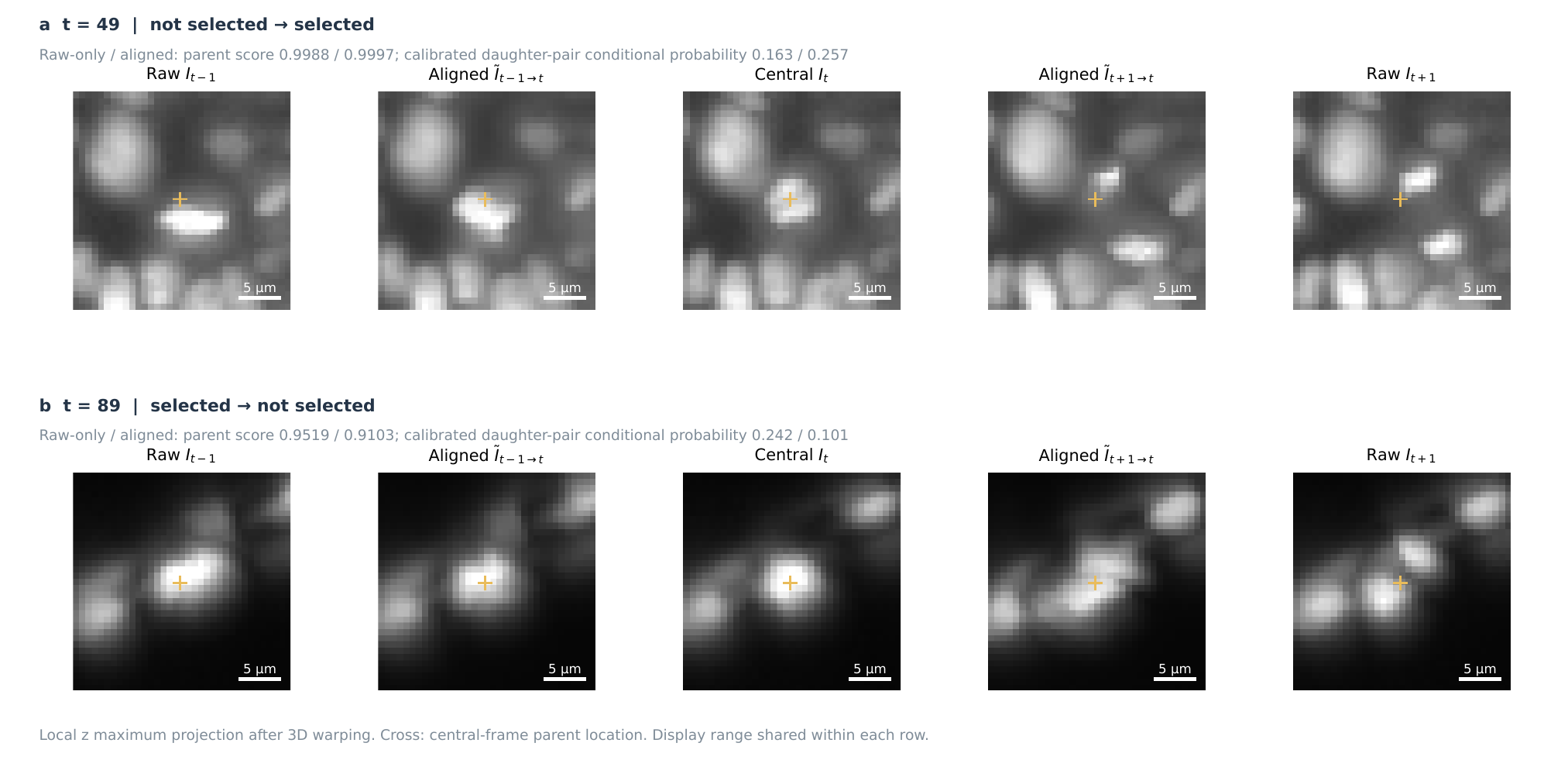}
\caption{\textbf{Aligned inputs and changes in division candidate selection.} Each row shows, from left to right, the raw preceding frame, its aligned version, the central frame, the aligned following frame, and the raw following frame. Maximum-intensity projections use the same local z range after 3D alignment and a shared intensity range within each row. Crosses mark the parent candidate's location in the central frame. Both examples are positive candidates corresponding to the two edges of a GT division. Replacing five raw-only inputs with five inputs including alignment changes the decision from not selected to selected in (a) and from selected to not selected in (b). The parent scores from both models and the recalibrated conditional probabilities of daughter-pair correctness are shown. Selection is determined by optimization with competing candidates and structural constraints, rather than by simple thresholding of these probabilities. These examples illustrate both directions of change in OOF outputs and do not estimate the frequency of improvement or differences in official division TP. Scale bars: 5 \ensuremath{\mu}m.}
\label{fig:alignment}
\end{figure*}

Alignment changes the score by +0.0022 after optimization and by +0.0023 in the final output (95\% CI [\ensuremath{-}0.0005, +0.0052]). In the final output, division FP decreases from 41 to 34, while adjusted edge Jaccard decreases slightly from 0.9235 to 0.9234. Although the point estimate of the overall score increases, the comparison does not establish a clear benefit from alignment.
Figure~\ref{fig:alignment} shows two candidate-level examples.

\subsection{Replacement with a Public Detector and Calibration Transfer}
\label{sec:transfer}

To evaluate transfer of the downstream models after changing the detector, we replace the detection stage with that from Pilkwang Kim's public notebook, Biohub Cell Tracking: Two Seeds Logit Blend~\cite{public-notebook}. This detector uses TemporalUNet3D to integrate temporal features from consecutive images. We use the notebook's publicly available primary-model and second-seed weights, rescale the second model's logits using the primary model's mean and standard deviation, and combine them by a weighted average with weight 0.475. Each model uses 8-way TTA, and the combined detection threshold is 0.96875. We compare direct transfer of the downstream models, retraining only the calibrators, and retraining the matcher, division models, and calibrators. Flow is shared across conditions.

The public detector was trained on all 199 training videos, so the following local evaluation includes overlap between detector training and evaluation data. We therefore use it as a diagnostic of downstream transfer under retraining, rather than as a detector comparison under the same CV conditions as Section~\ref{sec:evaluation}.
Table~\ref{tab:transfer} summarizes the results.

\begin{table*}[t]
\centering
\begin{threeparttable}
\caption{Downstream transfer and retraining with the public detector.}
\label{tab:transfer}
\normalsize
\setlength{\tabcolsep}{4pt}
\renewcommand{\arraystretch}{1.15}
\begin{tabular}{@{}lrrr@{}}
\toprule
Downstream configuration for the public detector & \shortstack[r]{Local diagnostic\\score} & Public & Private \\
\midrule
Transfer the original image models and calibrators & 0.9650 & 0.9523 & 0.9432 \\
Retrain calibrators only & 0.9669 & 0.9463 & 0.9319 \\
Retrain matcher, division models, and calibrators & 0.9676 & 0.9473 & 0.9357 \\
\bottomrule
\end{tabular}
\end{threeparttable}
\end{table*}

Retraining the calibrators increases the local diagnostic score from 0.9650 to 0.9669, while the private score decreases from 0.9432 to 0.9319. The local improvement does not carry over to the hidden test set, and downstream retraining does not improve transfer performance in this replacement experiment. Comparing the detectors' generalization performance would require training them under the same fold split.

\subsection{Discussion}
\label{sec:discussion}

\textbf{Interpretation of the results.} On fixed candidates, improvements are observed from joint selection of ordinary links and divisions, the calibration ensemble relative to a linear model, and auxiliary information after division. The final-output comparisons do not establish clear improvements from learned Exposure, an explicit node price, repeated price updates, auxiliary information before division, or flow-based input alignment.

\textbf{Detection and downstream tracking.} Cell detections are the shared input to correspondence estimation and division identification. Downstream classification alone cannot recover divisions with missing daughter detections or daughters outside the candidate search range. We evaluate the complete pipeline using the 2.5D and 3D detector ensemble trained for sparse annotations as a common foundation, and compare downstream components on fixed candidates. We do not individually ablate the detection mask, loss normalization, or ensemble members.

\textbf{Scope.} Because the official metric relies on annotated lineages, it does not comprehensively evaluate tracking accuracy in unannotated regions. Remaining challenges include the approximate inverse displacement used to align preceding frames, interpolation across long gaps, and the difference between division calibration targets and the temporal tolerance of the official evaluation.

\textbf{Computational design.} To reduce inference computation, the matcher computes only the intermediate features needed for correspondence estimation, and dense flow is shared between matching and division input alignment. Auxiliary division models using unaligned images can reuse encoder features across overlapping temporal windows.

\section{Conclusion}
\label{sec:conclusion}

We presented a 3D cell tracking method that integrates detection from sparse annotations, calibration using motion, appearance, and division features, and joint selection of ordinary links and divisions. The 2.5D and 3D models are trained with suppressed background supervision at unannotated cells, and their outputs are combined into a common set of detection candidates for downstream processing. Candidate connections are calibrated using features from dense flow, the matcher, and models that identify cells immediately before division, dividing parents, and daughters immediately after division. Ordinary links and divisions are selected jointly under lineage constraints. Gap filling and coordinate refinement based on temporal consistency further improve tracking performance.

The pipeline achieves an overall score of 0.9762 in 5-fold CV on all 199 videos. With candidates and calibrated values held fixed, joint selection outperforms a sequential method that alternates ordinary link and division optimization by 0.0105. Improvements are also observed for the calibration ensemble relative to a linear model and for the average effect of auxiliary information after division. These results support integrating motion, appearance, and division assessment of detection candidates with selection of connections across the lineage graph.

{\footnotesize
\bibliographystyle{ieee}
\bibliography{main}
}

\clearpage
\appendix
\numberwithin{equation}{section}
\section{Optimization Details}
\label{sec:optimization-details}

\subsection{Event Contributions and Count Priors}

For each event \(e\), define expected contributions \((\tau_e,\phi_e,\upsilon_e,\psi_e)\) to edge TP, edge FP, division TP, and division FP. An ordinary link has \((\tau_e,\phi_e)=(a_eq_e,a_e(1-q_e))\) and \(\upsilon_e=\psi_e=0\). For a division \(e=(i,\{j,k\})\), let \(\mathcal A_e=\{(i,j),(i,k)\}\). Then

\begin{equation}
\begin{aligned}
\tau_e&=\sum_{\ell\in\mathcal A_e}a_\ell q_\ell,&
\phi_e&=\sum_{\ell\in\mathcal A_e}a_\ell(1-q_\ell),\\
\upsilon_e&=a_e^{\mathrm{div}}q_e^{\mathrm{div}},&
\psi_e&=a_e^{\mathrm{div}}(1-q_e^{\mathrm{div}}).
\end{aligned}
\end{equation}

The basic division candidates reuse the two ordinary edges' calibrated values. In an expanded candidate, the representative edge retains its ordinary-edge values, and the added edge uses the expanded-edge calibrators. Thus a division's expected edge TP count is the sum of two edge probabilities, not twice the daughter-pair probability. Event-capacity constraints prevent simultaneous selection of events sharing the same physical edge.

For each video, \((T_s,F_s,U_s,V_s)\) is the sum of these four contributions weighted by \(x_e\). The node count is computed from the distinct endpoints of selected events, before gap filling. Let \(n_s^{\mathrm{cand}}\) be the initial number of detection candidates and \(\mathcal T_s\) the calibrator training videos for video \(s\). With training counts \(K_r,L_r,M_r\) for annotated edges, annotated divisions, and organizer-estimated total nodes, the priors are

\begin{equation}
(\widehat K_s,\widehat L_s,\widehat M_s)
=n_s^{\mathrm{cand}}
\frac{\sum_{r\in\mathcal T_s}(K_r,L_r,M_r)}
{\sum_{r\in\mathcal T_s}n_r^{\mathrm{cand}}}.
\end{equation}

These are ratios of pooled counts. For CV, \(\mathcal T_s\) excludes video \(s\)'s fold; hidden-test calibrators and priors use all training videos.

\subsection{Linearized Objective}

At the current feasible solution, write

\begin{equation}
\begin{aligned}
A&=\sum_s\max\{0,u_s(N_s)T_s\},&
B&=\sum_s(\widehat K_s+F_s),\\
C&=\sum_s U_s,&
D&=\sum_s(\widehat L_s+V_s).
\end{aligned}
\end{equation}

For positive denominators, let \(J_E=A/B\) and \(J_D=C/D\). Differentiation gives the edge TP reward \(r_{E,s}\), edge FP cost \(c_E\), division TP reward \(r_D\), and division FP cost \(c_D\):

\begin{equation}
\begin{aligned}
r_{E,s}&=\frac{\max\{0,u_s(N_s)\}}{B},&
c_E&=\frac{J_E}{B},\\
r_D&=\frac{0.1}{D},&
c_D&=\frac{0.1J_D}{D}.
\end{aligned}
\end{equation}

The node cost is \(c_{N,s}=0.1T_s/(\widehat M_s B)\) when \(0<1.1-0.1N_s/\widehat M_s<1\), and zero otherwise. This includes a zero-cost convention at both clipping boundaries. With all coefficients fixed, the linearized problem is

\begin{equation}
\begin{aligned}
\max_{x,y}\quad
&\sum_e\bigl(r_{E,s(e)}\tau_e-c_E\phi_e
+r_D\upsilon_e-c_D\psi_e\bigr)x_e\\
&-\sum_i c_{N,s(i)}y_i,
\end{aligned}
\end{equation}

subject to the constraints in Section~\ref{sec:optimization}. Here \(s(e)\) and \(s(i)\) identify the corresponding video. A zero-denominator surrogate term is defined as zero. For coefficient computation, \(B\) is floored at \(10^{-12}\), and the division coefficients are omitted when \(D=0\).

\subsection{Updates and Numerical Settings}

The control pipeline uses the following procedure, starting from the feasible initialization described in Section~\ref{sec:optimization}.

\begin{enumerate}
\item Pool expected counts across the videos being processed and compute the prices above. Solve each video's linearized problem independently using the common pooled quantities.
\item Reconstruct node inclusion from selected event endpoints, check feasibility, and pool the resulting expected counts. Accept the proposal only if its surrogate exceeds the incumbent's by more than \(10^{-10}\).
\item If the proposal is not accepted, update only \(J_D\) in the division FP cost to the proposed division ratio and solve again, keeping the incumbent, denominators, and other coefficients fixed. Allow at most three attempts per round, stopping the retries if the change in the division ratio is below \(10^{-10}\).
\item If all attempts fail, retain the incumbent and stop. Otherwise repeat from the accepted solution for at most three rounds. Post-processing is applied to the retained graph.
\end{enumerate}

Before each solve, objective coefficients are divided by their maximum absolute value, floored at \(10^{-12}\). The LP and MILP each have a 120-second time limit, and the MILP relative-gap target is zero. An LP solution is accepted as integral within \(10^{-6}\); MILP output is checked against an integrality tolerance of \(10^{-5}\). Rounded solutions must satisfy the full constraints within \(10^{-6}\). Zero-cost isolated nodes are removed before this check and before recomputing the surrogate. If the solver returns no solution, a feasible incumbent is retained; a returned solution that fails validation is rejected with an error. These checks concern feasibility and numerical termination, without establishing global optimality of the nonlinear surrogate.

\end{document}